\documentclass[letterpaper]{article} 
\usepackage{aaai2027}  
\nocopyright
\usepackage[hyphens]{url}  
\usepackage{graphicx} 
\usepackage{natbib}  
\usepackage{caption} 
\usepackage{algorithm}
\usepackage{algorithmic}

\usepackage{newfloat}
\usepackage{listings}
\DeclareCaptionStyle{ruled}{labelfont=normalfont,labelsep=colon,strut=off} 
\floatstyle{ruled}
\newfloat{listing}{tb}{lst}{}
\floatname{listing}{Listing}

\usepackage{booktabs}
\usepackage{colortbl}
\usepackage{pifont}
\definecolor{tablegroupgray}{RGB}{242,242,242}
\definecolor{oursrowblue}{RGB}{218,241,249}
\newcommand{\cmark}{\ding{51}}
\newcommand{\xmark}{\ding{55}}

\title{Auto-JEPA: A Latent World Model of Continuous Intent for End-to-End Autonomous Driving}
\author{
    Jiwei Yang\textsuperscript{\rm 1},
    Zhengxian Chen\textsuperscript{\rm 1,\rm 2}\textsuperscript{\textdagger},
    Chaosheng Huang\textsuperscript{\rm 1},
    Jun Li\textsuperscript{\rm 1}\\
}
\affiliations{
    \textsuperscript{\rm 1}School of Vehicle and Mobility, Tsinghua University\\
    \textsuperscript{\rm 2}State Key Laboratory of Intelligent Green Vehicle and Mobility, Tsinghua University\\
    chenzhengxian@tsinghua.vin
}

\begin{document}

\maketitle
\begingroup
\renewcommand{\thefootnote}{\textdagger}
\footnotetext{Corresponding author.}
\endgroup

\begin{abstract}
Existing autonomous-driving world models typically perform dense prediction of future videos, occupancy states, BEV representations, or agent motion. We argue that planning need not reconstruct the complete future world, but only focus on scene features that affect future ego action. Based on this perspective, we propose \textbf{Auto-JEPA}, an action-oriented latent world model that learns continuous future driving intent through joint-embedding prediction. Given visual observations, ego-motion history, and navigation commands, Auto-JEPA predicts an intent embedding aligned with the latent representation of the future ego trajectory. The predicted intent retrieves executable trajectories from a fixed trajectory memory, which are then ranked by a scene-conditioned candidate selection module. Auto-JEPA keeps the visual encoder frozen, requires no explicit perception annotations, and uses no learned trajectory generator. By optimizing only task-specific modules for trajectory representation, intent prediction, and candidate selection, Auto-JEPA achieves \textbf{91.3 PDMS} on NAVSIM v1 and \textbf{89.1 EPDMS} on NAVSIM v2. Semantic occlusion experiments show that masking dynamic-agent regions induces an average intent change $2.97\times$ that of equal-area random masking. Moreover, occluding vehicles that affect future driving substantially changes the predicted intent and selected trajectory, whereas both remain essentially unchanged when non-influential vehicles are occluded. These results show that future-intent prediction encourages the model to focus on planning-relevant visual features and supports high-quality planning without dense future-world modeling. \textbf{Code \& Models:} \url{https://github.com/NoctYang/Auto-JEPA}.
\end{abstract}


\section{Introduction}

\begin{figure}[t]
    \centering
    \includegraphics[width=\columnwidth]{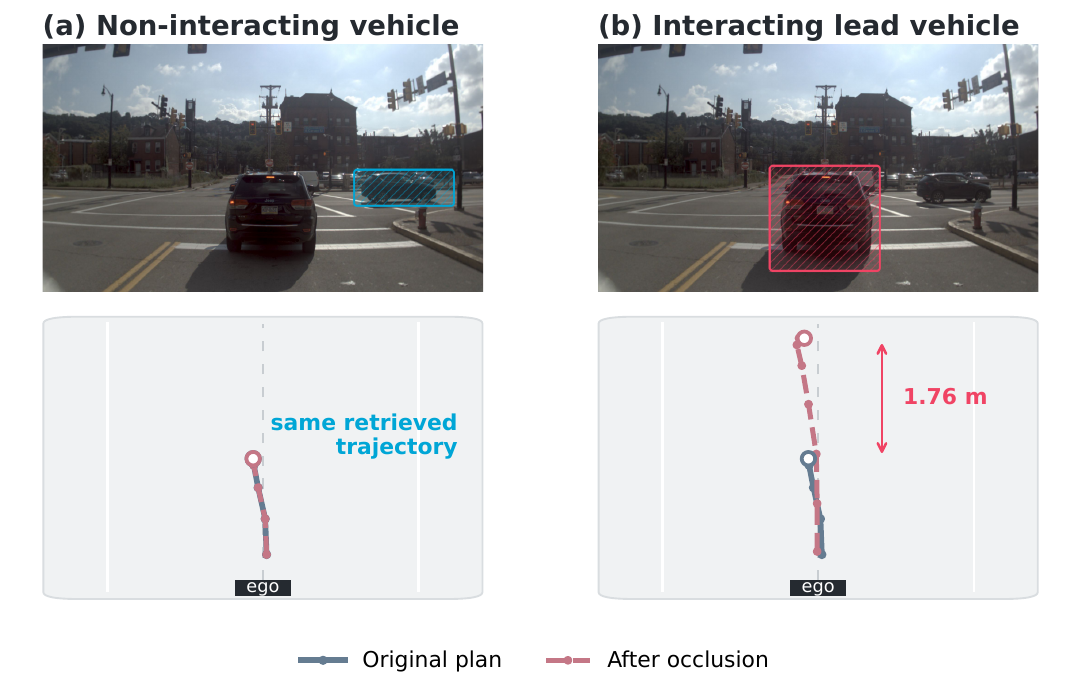}
    \caption{Selective response to action-relevant scene information. Occluding a non-interacting vehicle causes little change in the predicted intent and plan, whereas occluding the interacting lead vehicle shifts the intent and selected trajectory. Annotations are used only for analysis.}
    \label{fig:intro-selective-intent}
\end{figure}

World models offer a compelling route toward autonomous driving systems that can reason about how a scene may evolve before committing to an action. Many driving world models, however, formulate this objective as dense future prediction, generating future multiview observations, occupancy fields, or latent states that describe the evolution of the surrounding scene \cite{wang2023drivewm,zheng2023occworld,zhang2026deepsight,jia2026driveworldvla}. A parallel line of work explicitly forecasts the future trajectories and interactions of multiple traffic participants \cite{seff2023motionlm}. Although these objectives preserve rich information about the environment, they require the model to allocate substantial capacity to scene elements whose dynamics may have little influence on the ego vehicle's immediate decision. Predicting all observable entities without regard to their planning relevance can therefore increase computation and propagate perception and forecasting errors into downstream planning.

We argue that a planning-oriented predictive model need not reconstruct the complete future state of a scene, but only focus on scene features that affect future ego action. Based on this perspective, we introduce \textbf{Auto-JEPA}, an action-oriented latent world model that learns continuous future driving intent through joint-embedding prediction. Given visual observations, ego-motion history, and a navigation command, it predicts a future ego-trajectory latent whose temporal tokens encode motion geometry and dynamics. Recent work explores planning-aligned latent prediction \cite{li2024law,zheng2025world4drive,wang2026drivejepa,xie2026latentvla}, primarily for representation pretraining, auxiliary supervision, or action distillation. Auto-JEPA uses future ego-trajectory latents for trajectory retrieval.

Auto-JEPA first trains a trajectory encoder and then freezes it to define the future-trajectory target space. Following joint-embedding predictive learning \cite{assran2023ijepa,bardes2024vjepa}, a visual predictor infers the corresponding future-trajectory latent from the current scene context, ego-motion history, and route command. Latent alignment and contrastive objectives train the predictor without future-image reconstruction. Because its prediction target is defined solely by future ego motion, the model need not preserve all observable scene content and is encouraged to prioritize planning-relevant visual features. At inference, the predicted intent directly serves as the retrieval key of a fixed trajectory memory, and a scene-conditioned module selects the final executable candidate.

The trajectory memory contains recorded, kinematically plausible trajectory geometries encoded by the same trajectory encoder. A scene-conditioned scorer ranks the retrieved candidates, and a learned drivable-area gate screens candidates likely to violate drivable-area constraints before final selection. This design separates complementary responsibilities: the latent predictor determines \emph{what kind of future motion is appropriate}, retrieval provides explicit trajectory geometry, the scorer estimates scene-level driving quality, and the gate reduces drivable-area violations. Although these components are optimized in stages, Auto-JEPA preserves an end-to-end sensor-to-trajectory planning interface without intermediate perception outputs or a learned trajectory generator.

We evaluate Auto-JEPA on NAVSIM v1 and NAVSIM v2 \cite{dauner2024navsim}. Under this lightweight setting, Auto-JEPA achieves \textbf{91.3 PDMS} on NAVSIM v1 and \textbf{89.1 EPDMS} on NAVSIM v2. Figure~\ref{fig:intro-selective-intent} illustrates the different effects of individual vehicles within the same scene: occluding the interacting lead vehicle substantially changes the predicted intent and selected trajectory, whereas both remain essentially unchanged when the non-interacting adjacent vehicle is occluded. To quantify sensitivity to traffic-participant information at the dataset level, we further conduct controlled semantic occlusions on the full validation split. Masking dynamic-agent regions induces an average intent change $2.97\times$ that of equal-area random masking and a larger change in $71.1\%$ of scenes.

Our main contributions are summarized as follows:
\begin{itemize}
    \item We formulate planning-oriented latent world modeling as future ego-trajectory latent prediction. Using continuous driving intent as the prediction target encourages the model to focus on visual features relevant to ego action, thereby avoiding dense future-scene reconstruction.
    \item We propose Auto-JEPA, which maps visual context into a future-trajectory latent space through joint-embedding prediction. The predicted intent directly retrieves executable candidates from a fixed trajectory memory, followed by scene-conditioned selection.
    \item Auto-JEPA achieves \textbf{91.3 PDMS} on full NAVSIM v1 navtest and \textbf{89.1 EPDMS} on NAVSIM v2; controlled semantic occlusion experiments further show that the model selectively focuses on more critical scene features.
\end{itemize}

\section{Related Work}

\subsection{World Models for Autonomous Driving}

Driving world models predict scene evolution in pixel or structured spaces for simulation, representation learning, and planning. Generative approaches synthesize controllable future driving videos from observations, actions, or language \cite{hu2023gaia,wang2023drivedreamer,wang2023drivewm,gao2024vista}, while structured alternatives forecast future occupancy or point clouds \cite{zheng2023occworld,yang2023vidar}. Although these representations capture rich appearance and geometry, they require dense prediction over broad scene content, including details that need not determine the immediate ego plan.

Latent world models reduce explicit reconstruction by forecasting compact future features. LAW, World4Drive, DeepSight, and DriveWorld-VLA predict future scene features, BEV states, or action-conditioned scene evolution for downstream planning \cite{li2024law,zheng2025world4drive,zhang2026deepsight,jia2026driveworldvla}. Their targets nevertheless primarily describe how the surrounding scene evolves. Auto-JEPA instead predicts the representation of future ego motion itself, retaining scene dynamics through their implications for the planned trajectory.

\subsection{Joint-Embedding Predictive Learning}

Joint-embedding predictive architectures learn by predicting target representations rather than reconstructing observations. I-JEPA establishes this principle for images \cite{assran2023ijepa}, V-JEPA extends it to video \cite{bardes2024vjepa}, and V-JEPA~2 further demonstrates that self-supervised video representations can support understanding, prediction, and planning \cite{assran2025vjepa2}. Drive-JEPA adapts this family to driving-video pretraining and trajectory distillation \cite{wang2026drivejepa}. Auto-JEPA differs in the operational role of prediction: the future ego-trajectory latent directly serves as the planner's retrieval key and therefore participates in trajectory planning at inference rather than only supporting training-time representation learning.

\subsection{End-to-End Trajectory Planning}

End-to-end driving maps sensor observations and navigation context directly to an ego trajectory, but a single logged future provides limited coverage of multiple feasible maneuvers. Existing methods score candidates from offline trajectory vocabularies \cite{chen2024vadv2,li2024hydramdp,sun2026sparsedrivev2}, generate or refine proposals \cite{liao2025diffusiondrive,xing2025goalflow,guo2025ipad}, or combine a learned generator with a trajectory scorer \cite{ang2026clover}. Latent trajectory nearest-neighbor search has also been explored for motion forecasting rather than ego planning \cite{biktairov2020prank}.

Recent VLA planners tokenize or decode ego actions from visual and language context \cite{zhou2025autovla,zhou2025opendrivevla,li2025drivevlaw0}. They learn parametric action decoders, whereas Auto-JEPA predicts a continuous future-ego-motion intent latent for non-parametric trajectory retrieval.

\section{Method}

\subsection{Overview}

\begin{figure*}[t]
    \centering
    \includegraphics[width=\textwidth]{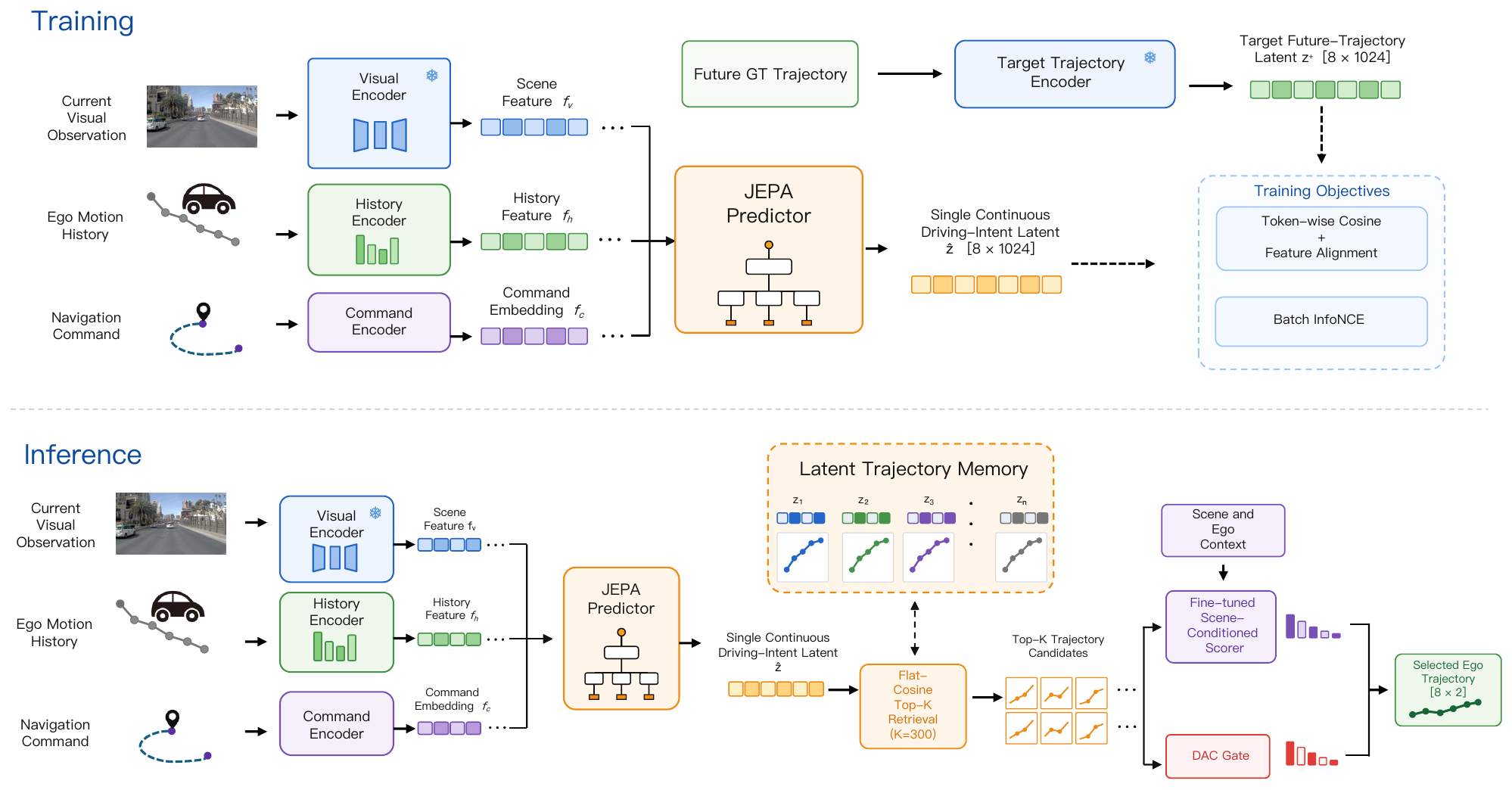}
    \caption{Overview of Auto-JEPA. During training, the predictor learns a continuous future ego-motion intent by aligning its predicted latent with the representation of the ground-truth future trajectory. During inference, the predicted intent retrieves 300 candidates from a ground-truth-only latent trajectory memory; a scene-conditioned scorer ranks candidate quality and an independent feasibility gate filters drivable-area violations before final selection. Snowflake symbols indicate frozen modules.}
    \label{fig:system-overview}
\end{figure*}

Auto-JEPA predicts a continuous future ego-motion representation and uses it to retrieve executable trajectories. Given four front-camera frames, four historical ego positions, and a route command, a frozen V-JEPA~2 encoder \cite{assran2025vjepa2} extracts visual tokens, and a Transformer predictor fuses visual, motion, and route context to produce eight temporal latent tokens representing one continuous driving intent. Figure~\ref{fig:system-overview} presents the complete training and inference pipeline.

During training, the predicted intent is aligned with the representation of the ground-truth future trajectory produced by a frozen trajectory encoder. At inference, it retrieves the 300 most similar trajectories from a non-parametric memory; a scene-conditioned scorer ranks these candidates and a learned drivable-area gate filters infeasible proposals. This separates intent prediction, trajectory instantiation, quality ranking, and feasibility filtering without reconstructing future observations or directly regressing waypoint coordinates.

\subsection{Future Ego-Motion Intent Representation}

\begin{figure}[t]
    \centering
    \includegraphics[width=\columnwidth]{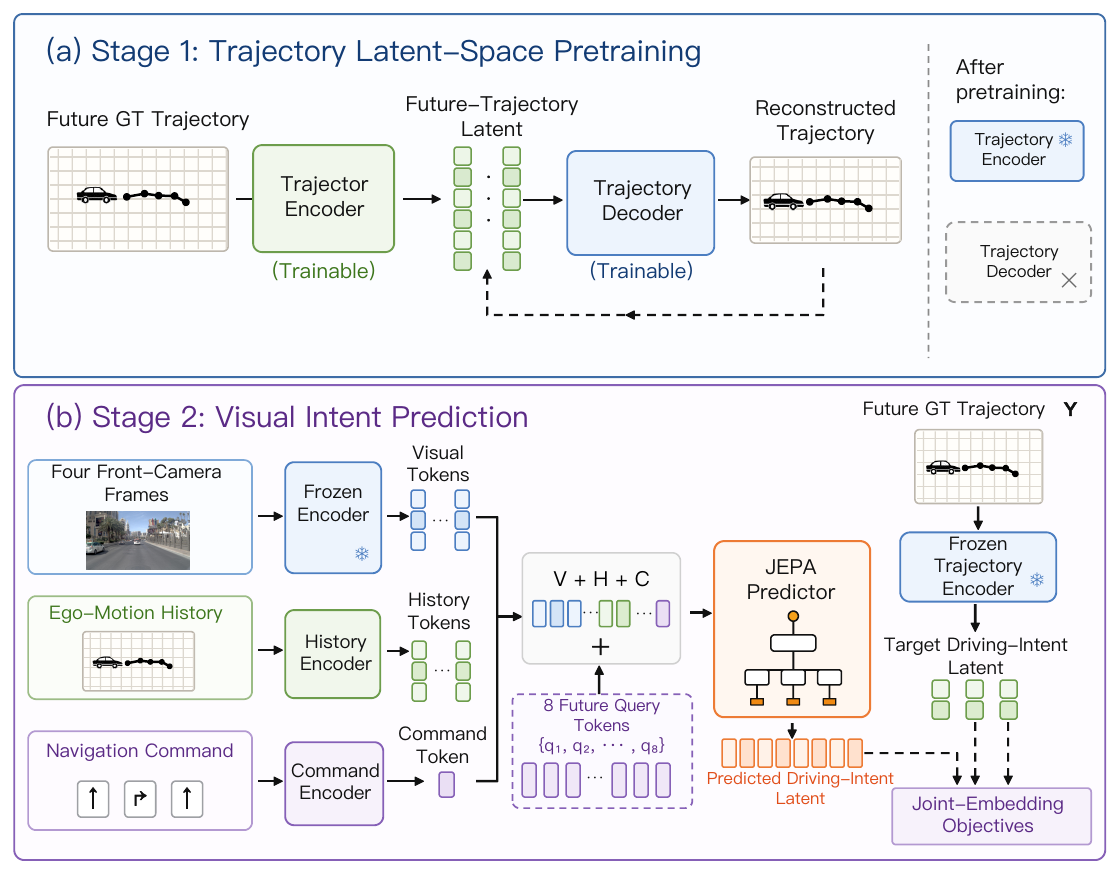}
    \caption{Trajectory-space pretraining and visual intent prediction in Auto-JEPA. Stage~1 learns the future-trajectory target space with a trajectory autoencoder; the decoder is then discarded and the trajectory encoder is frozen. Stage~2 predicts a continuous future driving-intent latent from visual observations, ego-motion history, and navigation commands, and aligns it with the frozen trajectory target representation.}
    \label{fig:method-overview}
\end{figure}

Figure~\ref{fig:method-overview} details the two-stage learning process: trajectory-space pretraining followed by visual intent prediction. For each driving scene, we represent the ground-truth future ego trajectory as eight two-dimensional waypoints,
\begin{equation}
    \mathbf{Y} = [(x_1,y_1),\ldots,(x_8,y_8)] \in \mathbf{R}^{8\times 2}.
\end{equation}
Before training the visual intent predictor, we learn the target space with a trajectory autoencoder composed of a trajectory encoder $E_{\mathrm{traj}}$ and a lightweight decoder $D_{\mathrm{traj}}$. The encoder maps a future trajectory to a sequence of latent tokens, and the decoder reconstructs the trajectory from this representation,
\begin{equation}
    \mathbf{Z}^{+} = E_{\mathrm{traj}}(\mathbf{Y})
    \in \mathbf{R}^{8\times 1024},
    \qquad
    \hat{\mathbf{Y}} = D_{\mathrm{traj}}(\mathbf{Z}^{+}).
\end{equation}
The autoencoder is optimized using a trajectory reconstruction objective,
\begin{equation}
    \mathcal{L}_{\mathrm{traj}}
    = \mathcal{L}_{\mathrm{xy}}
    + \lambda_{e}\mathcal{L}_{\mathrm{end}}
    + \lambda_{v}\mathcal{L}_{\mathrm{vel}}
    + \lambda_{a}\mathcal{L}_{\mathrm{acc}},
\end{equation}
where the four terms supervise waypoint coordinates, the final endpoint, velocity, and acceleration, respectively. After this pretraining stage, the decoder is discarded and $E_{\mathrm{traj}}$ is frozen. The frozen encoder then defines the target latent $\mathbf{Z}^{+}$ for intent prediction and encodes every trajectory in the retrieval memory, ensuring that prediction targets and retrieval candidates occupy the same latent space.

The eight latent tokens preserve trajectory time, geometry, and motion and jointly describe one continuous future realization rather than eight maneuver classes. We refer to this representation as the \emph{driving intent}. Using the same encoder for supervision and memory construction places predicted intents and executable trajectories in a shared space, allowing the intent latent to serve directly as the retrieval key.

\subsection{Visual Intent Prediction}

Given the current visual observation, ego-motion history, and navigation command, Auto-JEPA predicts the corresponding future ego-motion latent. We denote the model input as
\begin{equation}
    \mathbf{X} = (\mathbf{I},\mathbf{H},\mathbf{C}),
\end{equation}
where $\mathbf{I}$ contains four historical frames from the front-facing camera, $\mathbf{H}\in\mathbf{R}^{4\times2}$ contains four historical ego positions, and $\mathbf{C}\in\mathbf{R}^{4}$ denotes the route command. A frozen V-JEPA~2 visual encoder $E_{\mathrm{vis}}$ extracts visual tokens,
\begin{equation}
    \mathbf{F}_{v}=E_{\mathrm{vis}}(\mathbf{I}).
\end{equation}
The history encoder $E_{\mathrm{hist}}$ and command encoder $E_{\mathrm{cmd}}$ map the remaining inputs to the predictor feature dimension,
\begin{equation}
    \mathbf{F}_{h}=E_{\mathrm{hist}}(\mathbf{H}),
    \qquad
    \mathbf{F}_{c}=E_{\mathrm{cmd}}(\mathbf{C}).
\end{equation}

The visual, history, and command features condition a JEPA predictor $P_{\theta}$ composed of 24 Transformer blocks \cite{vaswani2017attention} with a hidden dimension of 1024 and 16 attention heads. The predictor outputs eight future latent tokens,
\begin{equation}
    \hat{\mathbf{Z}}
    =P_{\theta}(\mathbf{F}_{v},\mathbf{F}_{h},\mathbf{F}_{c})
    \in\mathbf{R}^{8\times1024}.
\end{equation}
The prediction $\hat{\mathbf{Z}}$ matches the temporal organization and feature dimension of $\mathbf{Z}^{+}$. During driving-domain training, the visual encoder remains frozen, while the history encoder, command encoder, and JEPA predictor are optimized to aggregate visual evidence, ego dynamics, and navigation constraints without explicitly predicting intermediate scene states.

\subsection{Joint-Embedding Training Objectives}

Rather than directly supervising waypoint coordinates with ADE or FDE, Auto-JEPA optimizes predictions in the frozen trajectory latent space using feature alignment, token-wise cosine alignment, and batch-level InfoNCE \cite{oord2018cpc}.

\paragraph{Feature alignment.}
We first normalize the predicted and target latents and apply a Smooth L1 loss to their feature values,
\begin{equation}
    \mathcal{L}_{\mathrm{feat}}
    =\mathrm{SmoothL1}\!\left(
    \mathrm{Norm}(\hat{\mathbf{Z}}),
    \mathrm{Norm}(\mathbf{Z}^{+})
    \right).
\end{equation}
\paragraph{Token-wise cosine alignment.}
For the eight future time positions, we minimize the cosine distance between corresponding predicted and target tokens,
\begin{equation}
    \mathcal{L}_{\mathrm{cos}}
    =\frac{1}{8}\sum_{t=1}^{8}\left(
    1-\frac{\hat{\mathbf{z}}_{t}^{\top}\mathbf{z}^{+}_{t}}
    {\|\hat{\mathbf{z}}_{t}\|_{2}\|\mathbf{z}^{+}_{t}\|_{2}}
    \right).
\end{equation}
\paragraph{Batch-level InfoNCE.}
Positive alignment alone may map distinct driving scenes to similar representations. We therefore flatten and normalize each complete latent sequence,
\begin{equation}
    \hat{\mathbf{q}}_{i}
    =\mathrm{Norm}(\mathrm{vec}(\hat{\mathbf{Z}}_{i})),
    \qquad
    \mathbf{k}_{j}
    =\mathrm{Norm}(\mathrm{vec}(\mathbf{Z}^{+}_{j})).
\end{equation}
For scene $i$, its target latent is the positive and target latents from other scenes serve as negatives,
\begin{equation}
    \mathcal{L}_{\mathrm{NCE}}
    =-\frac{1}{B}\sum_{i=1}^{B}
    \log\frac{\exp(\hat{\mathbf{q}}_{i}^{\top}\mathbf{k}_{i}/\tau)}
    {\sum_{j=1}^{B}\exp(\hat{\mathbf{q}}_{i}^{\top}\mathbf{k}_{j}/\tau)},
\end{equation}
where $\tau=0.07$. During distributed training, target latents are gathered across GPUs to enlarge the negative set.

The complete intent-prediction objective is
\begin{equation}
    \mathcal{L}_{\mathrm{intent}}
    =0.1\mathcal{L}_{\mathrm{feat}}
    +2.0\mathcal{L}_{\mathrm{cos}}
    +\mathcal{L}_{\mathrm{NCE}}.
\end{equation}
Together, these objectives align local features and temporal-token semantics while preserving discrimination across driving scenes, making the predicted latent suitable for trajectory retrieval.

\subsection{Non-Parametric Trajectory Retrieval}

To ground the predicted continuous intent into explicit trajectory geometry, we construct a non-parametric trajectory memory using only ground-truth driving trajectories. Each memory trajectory $\mathbf{Y}_{n}$ is encoded by the same frozen trajectory encoder used to define the training targets,
\begin{equation}
    \mathbf{Z}_{n}=E_{\mathrm{traj}}(\mathbf{Y}_{n})
    \in\mathbf{R}^{8\times1024}.
\end{equation}
The resulting memory is
\begin{equation}
    \mathcal{M}
    =\left\{(\mathbf{Z}_{n},\mathbf{Y}_{n})\right\}_{n=1}^{N},
    \qquad N=110{,}335.
\end{equation}
Each entry retains its latent and waypoint coordinates, directly providing candidate geometry.

For a predicted intent $\hat{\mathbf{Z}}$, we flatten and L2-normalize the query and memory latents,
\begin{equation}
    \mathbf{q}=\mathrm{Norm}(\mathrm{vec}(\hat{\mathbf{Z}})),
    \qquad
    \mathbf{m}_{n}=\mathrm{Norm}(\mathrm{vec}(\mathbf{Z}_{n})).
\end{equation}
Their retrieval similarity is the flat cosine similarity
\begin{equation}
    r_{n}=\mathbf{q}^{\top}\mathbf{m}_{n}.
\end{equation}
We rank the complete memory by $r_n$ and retrieve the $K=300$ most similar trajectories,
\begin{equation}
    \mathcal{C}
    =\mathrm{TopK}\!\left(\{r_n\}_{n=1}^{N},K\right).
\end{equation}
Retrieval identifies intent-compatible geometry, while subsequent modules handle scene-level safety. Because candidates come from recorded trajectories, inference requires neither an additional learned trajectory generator nor iterative waypoint sampling.

\newcommand{\experimentfloats}{%
\begin{table*}[!t]
    \centering
    \small
    \setlength{\tabcolsep}{5.5pt}
    \caption{Comparison with representative methods on NAVSIM v1 navtest. C and L denote camera and LiDAR input, respectively. NC, DAC, TTC, C, EP, and PDMS denote no-at-fault collision, drivable-area compliance, time to collision, comfort, ego progress, and the Predictive Driver Model Score.}
    \label{tab:navsim-v1}
    \resizebox{\textwidth}{!}{%
    \begin{tabular}{ll|l|rrrrrr}
        \toprule
        Method & Venue & Sensors & NC $\uparrow$ & DAC $\uparrow$ & TTC $\uparrow$ & C $\uparrow$ & EP $\uparrow$ & PDMS $\uparrow$ \\
        \midrule
        Human & -- & -- & 100.0 & 100.0 & 100.0 & 99.9 & 87.5 & 94.8 \\
        \midrule
        \rowcolor{tablegroupgray}\multicolumn{9}{l}{\emph{End-to-End Planning Methods}} \\
        TransFuser \cite{chitta2023transfuser} & TPAMI 2023 & $3\times$C+L & 97.7 & 92.8 & 92.8 & \textbf{100.0} & 79.2 & 84.0 \\
        PARA-Drive \cite{weng2024paradrive} & CVPR 2024 & $6\times$C & 97.9 & 92.4 & 93.0 & 99.8 & 79.3 & 84.0 \\
        Hydra-MDP \cite{li2024hydramdp} & CVPR 2024 & $3\times$C+L & 98.3 & 96.0 & 94.6 & \textbf{100.0} & 78.7 & 86.5 \\
        DiffusionDrive \cite{liao2025diffusiondrive} & CVPR 2025 & $3\times$C+L & 98.2 & 96.2 & 94.7 & \textbf{100.0} & 82.2 & 88.1 \\
        \midrule
        \rowcolor{tablegroupgray}\multicolumn{9}{l}{\emph{World-Model-Based Methods}} \\
        LAW \cite{li2024law} & ICLR 2025 & $1\times$C & 96.4 & 95.4 & 88.7 & 99.9 & 81.7 & 84.6 \\
        DrivingGPT \cite{chen2025drivinggpt} & ICCV 2025 & $1\times$C & 98.9 & 90.7 & 94.9 & 95.6 & 79.7 & 82.4 \\
        WoTE \cite{li2025wote} & ICCV 2025 & $3\times$C+L & 98.5 & 96.8 & 94.4 & 99.9 & 81.9 & 88.3 \\
        Epona \cite{zhang2025epona} & ICCV 2025 & $3\times$C & 97.9 & 95.1 & 93.8 & 99.9 & 80.4 & 86.2 \\
        \midrule
        \rowcolor{tablegroupgray}\multicolumn{9}{l}{\emph{VLA-Based Methods}} \\
        AutoVLA \cite{zhou2025autovla} & NeurIPS 2025 & $3\times$C & 98.4 & 95.6 & \textbf{98.0} & 99.9 & 81.9 & 89.1 \\
        RecogDrive \cite{li2025recogdrive} & ICLR 2026 & $3\times$C & 98.2 & 97.8 & 95.2 & 99.8 & 83.5 & 89.6 \\
        AdaThinkDrive \cite{luo2025adathinkdrive} & ICRA 2026 & $1\times$C & 98.4 & 97.8 & 95.2 & \textbf{100.0} & 84.4 & 90.3 \\
        DriveVLA-W0 \cite{li2025drivevlaw0} & ICLR 2026 & $1\times$C & 98.7 & \textbf{99.1} & 95.3 & 99.3 & 83.3 & 90.2 \\
        Curious-VLA \cite{chen2026curiousvla} & CVPR 2026 Findings & $1\times$C & 98.4 & 96.9 & 97.9 & 98.1 & \textbf{88.5} & 90.3 \\
        \midrule
        \rowcolor{oursrowblue}\textbf{Auto-JEPA (Ours)} & -- & $1\times$C & 98.4 & 98.3 & 95.0 & \textbf{100.0} & 87.1 & \textbf{91.3} \\
        \bottomrule
    \end{tabular}%
    }
\par\vspace{6pt}
    \setlength{\tabcolsep}{4.2pt}
    \caption{Comparison with representative methods on NAVSIM v2. When multiple backbones are reported, we use the strongest source-reported configuration. For Auto-JEPA, the unmarked result uses the original evaluation implementation, while the result marked with $^\dagger$ uses the updated official implementation. Results for other methods are source-reported.}
    \label{tab:navsim-v2}
    \resizebox{\textwidth}{!}{%
    \begin{tabular}{l|rrrrrrrrrr}
        \toprule
        Method & NC $\uparrow$ & DAC $\uparrow$ & DDC $\uparrow$ & TL $\uparrow$ & EP $\uparrow$ & TTC $\uparrow$ & LK $\uparrow$ & HC $\uparrow$ & EC $\uparrow$ & EPDMS $\uparrow$ \\
        \midrule
        \rowcolor{tablegroupgray}\multicolumn{11}{l}{\emph{End-to-End Planning Methods}} \\
        TransFuser \cite{chitta2023transfuser} & 96.9 & 89.9 & 97.8 & 99.7 & 87.1 & 95.4 & 92.7 & 98.3 & 87.2 & 76.7 \\
        VADv2 \cite{chen2024vadv2} & 97.3 & 91.7 & 98.2 & \textbf{99.9} & 77.6 & 92.7 & 66.0 & \textbf{100.0} & \textbf{97.4} & 76.6 \\
        DiffusionDrive \cite{liao2025diffusiondrive} & 98.2 & 95.9 & 99.4 & 99.8 & 87.5 & 97.3 & 96.8 & 98.3 & 87.7 & 84.5 \\
        HydraMDP++ (ViT-L) \cite{li2025hydramdpp} & 98.5 & 98.5 & 99.5 & 99.7 & 87.4 & 97.9 & 95.8 & 98.2 & 75.7 & 85.6 \\
        DriveSuprim (ViT-L) \cite{yao2025drivesuprim} & 98.4 & 98.6 & \textbf{99.6} & 99.8 & \textbf{90.5} & 97.8 & \textbf{97.0} & 98.3 & 78.6 & 87.1 \\
        \midrule
        \rowcolor{tablegroupgray}\multicolumn{11}{l}{\emph{VLA-Based Methods}} \\
        DriveVLA-W0 \cite{li2025drivevlaw0} & 98.5 & \textbf{99.1} & 98.0 & 99.7 & 86.4 & \textbf{98.1} & 93.2 & 97.9 & 58.9 & 86.1 \\
        ReCogDrive \cite{li2025recogdrive} & 98.3 & 95.2 & 99.5 & 99.8 & 87.1 & 97.5 & 96.6 & 98.3 & 86.5 & 83.6 \\
        Curious-VLA \cite{chen2026curiousvla} & 98.4 & 96.9 & 99.2 & 99.8 & 88.5 & 97.9 & 96.9 & 98.1 & 81.5 & 85.3 \\
        DriveWorld-VLA \cite{jia2026driveworldvla} & \textbf{98.6} & \textbf{99.1} & \textbf{99.6} & 99.8 & 87.4 & 97.9 & \textbf{97.0} & 97.8 & 78.6 & 86.8 \\
        \midrule
        \rowcolor{oursrowblue}\textbf{Auto-JEPA (Ours)} & 98.5 & 98.7 & 98.2 & 97.2 & \textbf{90.5} & 97.9 & 84.0 & 97.8 & 75.4 & 85.6 \\
        \rowcolor{oursrowblue}\textbf{Auto-JEPA (Ours)$^\dagger$} & 98.5 & 98.7 & 98.3 & 99.7 & 90.5 & 97.9 & 94.7 & 97.8 & 75.2 & \textbf{89.1} \\
        \bottomrule
    \end{tabular}%
    }
\end{table*}

\begin{figure*}[t]
    \centering
    \includegraphics[width=\textwidth]{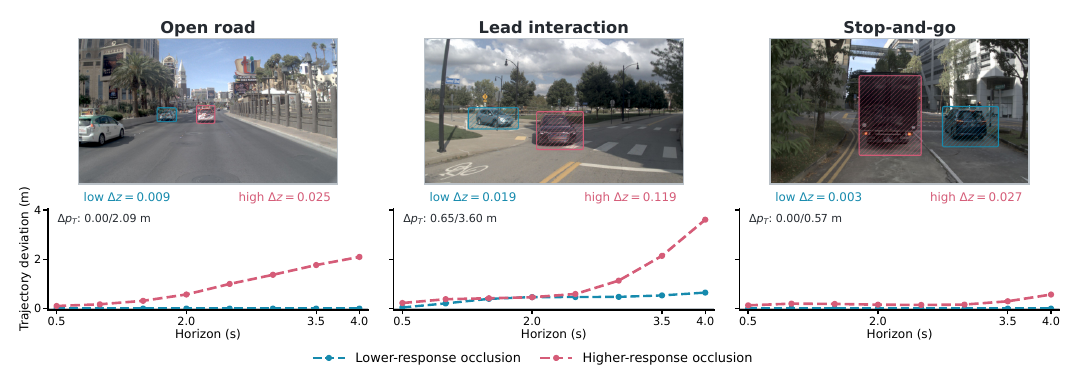}
    \caption{Selective responses to traffic participants. Cyan and rose denote occlusions of lower- and higher-impact vehicles, respectively. Curves show deviation from the unoccluded trajectory on a shared 0--4\,m scale; $\Delta p_T$ lists their terminal deviations in the same order. The smaller stop-and-go shift reflects its restricted motion range: the unoccluded plan moves only 0.17\,m.}
    \label{fig:semantic-occlusion}
\end{figure*}

\begin{figure}[t]
    \centering
    \includegraphics[width=0.94\columnwidth]{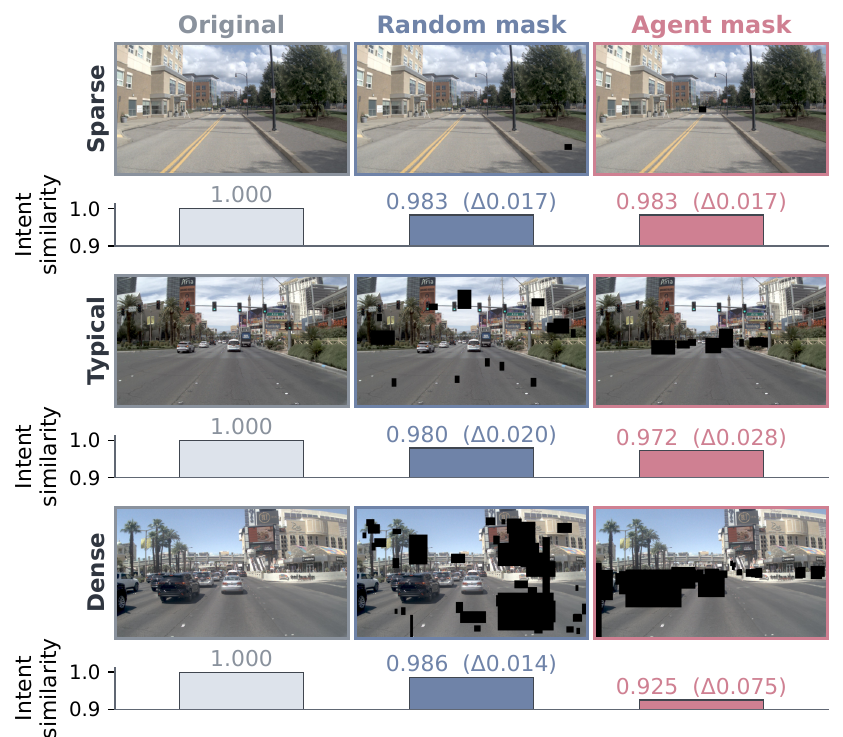}
    \caption{Representative controls from the full-validation semantic occlusion protocol. For each scene, dynamic-agent regions and independently sampled equal-area random regions are masked consistently across all four input frames. The bars report cosine similarity to the unoccluded intent.}
    \label{fig:semantic-occlusion-controls}
\end{figure}
}

\subsection{Scene Scoring and Feasibility Gating}

Latent similarity measures intent compatibility but not scene-level safety. We therefore use a scene-conditioned quality scorer and an independently trained drivable-area feasibility gate.

\paragraph{Scene-conditioned utility branch.}
Given scene features $\mathbf{F}_{\mathrm{scene}}$, ego context $\mathbf{e}$, and a candidate trajectory $\mathbf{Y}_{k}$, the scorer predicts a quality score
\begin{equation}
    s_k
    =S_{\phi}(\mathbf{F}_{\mathrm{scene}},\mathbf{e},\mathbf{Y}_{k}).
\end{equation}
We initialize $S_{\phi}$ from the publicly released CLOVER trajectory scorer \cite{ang2026clover} and re-optimize its trainable modules on the ground-truth-only candidates retrieved by Auto-JEPA. Training uses collision, drivable-area, time-to-collision, comfort, and ego-progress supervision \cite{dauner2024navsim}, together with a within-scene ranking objective. The scorer and gate are trained exclusively on candidates from the NAVSIM training split, using labels generated offline by the NAVSIM/CLOVER \texttt{get\_sub\_score} evaluator with the batched \texttt{navsim\_v1\_style} relabeling protocol. This protocol disables per-proposal two-way rollout during label generation; the final benchmark results are evaluated separately under the official NAVSIM protocols. Scorer training is separate from intent prediction, with no gradients propagated to the visual encoder, JEPA predictor, or trajectory memory.

\paragraph{Drivable-area feasibility gate.}
To explicitly reject candidates at risk of leaving the drivable region, the independently trained feasibility gate predicts
\begin{equation}
    p^{\mathrm{DAC}}_k
    =G_{\psi}(\mathbf{F}_{\mathrm{scene}},\mathbf{e},\mathbf{Y}_{k}),
\end{equation}
where $p^{\mathrm{DAC}}_k$ is the predicted probability of a DAC failure. At inference, we form a safety mask using $\tau_{\mathrm{DAC}}=0.2$,
\begin{equation}
    m_k
    =\mathbf{1}\!\left[p^{\mathrm{DAC}}_k\leq\tau_{\mathrm{DAC}}\right].
\end{equation}
Only candidates that pass the gate participate in final selection. If all are rejected, the system falls back to ungated utility ranking. Evaluator labels are unavailable to the gate at inference time.

The final trajectory is selected by a masked argmax over utility scores,
\begin{equation}
    k^{*}=\arg\max_{k:m_k=1}s_k,
    \qquad
    \mathbf{Y}^{*}=\mathbf{Y}_{k^{*}}.
\end{equation}
Thus, intent prediction supplies the retrieval query, memory provides trajectory geometry, and the scorer--gate cascade performs scene-conditioned selection.

\experimentfloats

\section{Experiments}

We evaluate the complete retrieval-based planner under the official NAVSIM v1 and v2 protocols. Unless otherwise stated, all experiments use the same intent predictor, ground-truth-only trajectory memory, scene-conditioned scorer, and feasibility gate. We first report benchmark results, then isolate the effects of intent prediction and candidate selection, followed by analyses of candidate-pool size and visual dependence.

\subsection{Dataset and Metrics}
\paragraph{NAVSIM v1.}
NAVSIM is a data-driven, non-reactive benchmark for autonomous-driving planning that provides large-scale trainval data and simulation-based evaluation on navtest \cite{dauner2024navsim}. The v1 benchmark evaluates each predicted ego trajectory through a simulator and reports no-at-fault collision (NC), drivable-area compliance (DAC), time to collision (TTC), comfort (C), and ego progress (EP). These terms are aggregated into the Predictive Driver Model Score (PDMS):
\begin{equation}
    \mathrm{PDMS}=\mathrm{NC}\,\mathrm{DAC}\,\frac{5(\mathrm{EP}+\mathrm{TTC})+2\mathrm{C}}{12}.
\end{equation}
We train on NAVSIM trainval and report results on the complete navtest split of 12,146 scenarios.

\paragraph{NAVSIM v2.}
NAVSIM v2 extends the evaluation to a broader set of driving-quality and rule-compliance properties. In addition to NC and DAC, it measures driving-direction compliance (DDC), traffic-light compliance (TL), ego progress (EP), time to collision (TTC), lane keeping (LK), history comfort (HC), and extended comfort (EC), and aggregates them into EPDMS. We evaluate v2 with the updated official implementation and human-behavior filtering enabled. Since v1 and v2 use different rollout and aggregation protocols, we report them separately without converting scores between the two metrics.

\subsection{Implementation Details}
The model takes four $256\times256$ front-camera frames, four historical ego positions, and a route command. Its 24-layer Transformer predictor has 16 heads and a hidden dimension of 1024, and outputs eight temporal latent tokens. We train with a per-GPU batch size of 8, learning rate $10^{-5}$, weight decay $0.05$, BF16 arithmetic, and an InfoNCE temperature of 0.07. The visual and target trajectory encoders remain frozen during predictor training. No object boxes, occupancy, semantic-map, or surrounding-agent motion labels are used. The scorer and gate use 75,823 training scenes and 8,459 validation scenes under a scene-prefix split; the validation and navtest sets have zero overlap in metric tokens. At inference, flat-cosine retrieval selects 300 candidates from a memory of 110,335 ground-truth trajectory--latent pairs; the scene scorer and feasibility gate then select the final trajectory using a DAC-failure threshold of 0.2.

\subsection{Results on NAVSIM v1}

As shown in Table~\ref{tab:navsim-v1}, Auto-JEPA achieves \textbf{91.3 PDMS} using only a front camera, with 98.4 NC, 98.3 DAC, and 100.0 comfort.

\subsection{Results on NAVSIM v2}

Table~\ref{tab:navsim-v2} reports 85.6 EPDMS under the original evaluator and \textbf{89.1 EPDMS} under the updated official implementation with human-behavior filtering. CLOVER reports 90.4 EPDMS with a learned generator--scorer pipeline; Auto-JEPA remains competitive without parametric proposal generation. The evaluator change mainly affects TL and LK.

\subsection{Ablation Studies}

\paragraph{Component ablation.}
We ablate each component using the same Top-300 memory. Replacing the predicted intent with a fixed codebook medoid reduces PDMS from 91.3 to 52.6, confirming scene-conditioned retrieval. Intent-based cosine retrieval with the gate already reaches 87.6; adding the scorer contributes 3.7 points. Removing the gate yields 91.0 PDMS and lowers DAC from 98.3 to 97.9.

\begin{table}[t]
    \centering
    \small
    \setlength{\tabcolsep}{2.5pt}
    \caption{Component ablation on full NAVSIM v1 navtest. Checkmarks indicate enabled components.}
    \label{tab:candidate-selection}
    \begin{tabular}{cccrrrrrr}
        \toprule
        Intent & Scorer & Gate & NC & DAC & TTC & C & EP & PDMS \\
        \midrule
        \xmark & \cmark & \cmark & 83.1 & 85.3 & 76.2 & 86.3 & 37.8 & 52.6 \\
        \cmark & \xmark & \cmark & 98.1 & 96.4 & 94.0 & \textbf{100.0} & 81.7 & 87.6 \\
        \cmark & \cmark & \xmark & \textbf{98.5} & 97.9 & \textbf{95.0} & 99.9 & 86.9 & 91.0 \\
        \cmark & \cmark & \cmark & 98.4 & \textbf{98.3} & \textbf{95.0} & \textbf{100.0} & \textbf{87.1} & \textbf{91.3} \\
        \bottomrule
    \end{tabular}
\end{table}

\paragraph{Candidate-pool size sensitivity.}
The system obtains 87.6, 91.1, and 91.3 PDMS for $K=1$, 200, and 300. Increasing $K$ from 1 to 200 improves PDMS by 3.5 points, demonstrating the value of candidate selection, whereas the marginal gain from 200 to 300 indicates that performance is approaching saturation.

\subsection{Analysis}
\label{sec:analysis}

\paragraph{Selective sensitivity to dynamic-agent information.}
Across 15{,}364 validation samples, masking all dynamic agents in all four frames produces a mean intent change ($1-$cosine similarity) of 0.080, compared with 0.027 for independently sampled equal-area masks, a $2.97\times$ increase; the dynamic-agent intervention is larger in 71.1\% of samples. Figure~\ref{fig:semantic-occlusion-controls} shows matched controls, while Figure~\ref{fig:semantic-occlusion} isolates individual vehicles. Vehicles affecting future driving cause larger latent and trajectory changes than low-impact vehicles. Differences grow with the horizon in the open-road and lead-interaction scenes, whereas the smaller stop-and-go response reflects an unoccluded plan that advances only 0.17\,m. Although the predictor receives no object boxes, agent identities, interaction labels, or surrounding-agent motion annotations, these responses emerge from future ego-trajectory representation supervision. These results show that the model does not respond uniformly to all dynamic agents, but focuses more strongly on vehicles that may affect future driving decisions.

\section{Conclusion}
This paper introduced Auto-JEPA, an action-oriented latent world model that predicts continuous future driving intent. Joint-embedding prediction maps visual observations, ego-motion history, and navigation commands into the latent space of future ego trajectories; the predicted intent then retrieves executable candidates from a fixed memory for scene-conditioned selection. Auto-JEPA achieves 91.3 PDMS on NAVSIM v1 and 89.1 EPDMS on NAVSIM v2. Ablations and controlled occlusions validate the model's selective focus on scene features, supporting future-trajectory latent as a decision-relevant prediction target without dense future-world modeling. The current system remains limited by memory coverage and selection calibration; intent-conditioned trajectory generation or refinement provides a natural extension.

\FloatBarrier

\bibliography{aaai2027,intent_jepa}

\clearpage
\appendix
\section*{Supplementary Material}
This supplementary material provides additional implementation details, training protocols, and analysis settings for Auto-JEPA. The organization follows the staged training and inference pipeline described in the main paper.

\section{Additional Implementation Details}

\subsection{Trajectory Latent-Space Pretraining}

The trajectory representation is learned before visual intent prediction. Each future ego trajectory is represented by eight planar waypoints,
\begin{equation}
    \mathbf{Y}\in\mathbf{R}^{8\times2},
\end{equation}
covering a four-second planning horizon at 0.5-second intervals. Coordinates are normalized by a scale factor of 64 before being processed by the trajectory autoencoder. The trajectory encoder contains four Transformer blocks with a hidden dimension of 1024, 16 attention heads, an MLP ratio of 4, and eight Fourier frequency bands for coordinate encoding. It produces eight temporally aligned latent tokens,
\begin{equation}
    \mathbf{Z}^{+}=E_{\mathrm{traj}}(\mathbf{Y})\in\mathbf{R}^{8\times1024}.
\end{equation}

The lightweight decoder contains four self-attention blocks with the same hidden dimension and number of attention heads. It predicts waypoint increments, which are cumulatively summed to obtain the reconstructed trajectory. The autoencoder is optimized with coordinate, endpoint, velocity, and acceleration consistency losses:
\begin{equation}
    \mathcal{L}_{\mathrm{traj}}
    =\mathcal{L}_{xy}
    +2.0\mathcal{L}_{\mathrm{end}}
    +0.5\mathcal{L}_{\mathrm{vel}}
    +0.2\mathcal{L}_{\mathrm{acc}}.
\end{equation}
We use a batch size of 256, a learning rate of $2\times10^{-4}$, weight decay of 0.05, dropout of 0.1, gradient clipping at 1.0, and BF16 arithmetic. The model is trained for up to 40 epochs, and the checkpoint with the lowest validation ADE is retained. After this stage, the decoder is discarded and the trajectory encoder is frozen for both intent supervision and trajectory-memory construction.

\subsection{Visual Intent Predictor}

The visual intent predictor receives four front-camera frames, four historical ego positions, and a route command. The input images are resized to $256\times256$ and encoded by a frozen V-JEPA~2 visual encoder. The history and route inputs are projected into the same 1024-dimensional feature space. A 24-layer Transformer predictor with 16 attention heads fuses these inputs with eight learnable future-time query tokens and predicts
\begin{equation}
    \hat{\mathbf{Z}}\in\mathbf{R}^{8\times1024}.
\end{equation}
The eight output tokens correspond to future time steps and jointly represent a single continuous driving intent; they are not separate maneuver queries.

The predictor is trained against the frozen target representation $\mathbf{Z}^{+}$ using feature alignment, token-wise cosine alignment, and batch-level contrastive learning:
\begin{equation}
    \mathcal{L}_{\mathrm{intent}}
    =0.1\mathcal{L}_{\mathrm{feat}}
    +2.0\mathcal{L}_{\mathrm{cos}}
    +\mathcal{L}_{\mathrm{NCE}}.
\end{equation}
The InfoNCE loss uses flattened trajectory latents and a temperature of 0.07. Training uses a per-GPU batch size of 8, learning rate $10^{-5}$, weight decay of 0.05, dropout of 0.1, BF16 arithmetic, and gradient clipping at 1.0. The image augmentation applies temporal frame masking with probability 0.3, masking between one and three input frames, and random image erasing with probability 0.2. When an image frame is masked, the corresponding historical ego position is masked as well.

\subsection{Trajectory Memory and Retrieval}

The trajectory memory contains 110,335 ground-truth trajectory--latent pairs constructed from the NAVSIM training data. Every trajectory is encoded once with the frozen trajectory encoder. At inference, the predicted intent and memory latents are flattened and $\ell_2$ normalized, and retrieval is performed by flat cosine similarity. The final configuration retrieves the Top-300 candidates. The memory is fixed during planner training and inference, and the NAVSIM navtest scenes are not included in memory construction.

\section{Dataset and Evaluation Protocol}

\subsection{NAVSIM Benchmarks}

NAVSIM provides a data-driven, non-reactive benchmark for autonomous-driving planning. NAVSIM v1 evaluates a predicted ego trajectory using no-at-fault collision, drivable-area compliance, time to collision, comfort, and ego progress, which are aggregated into PDMS. NAVSIM v2 extends the evaluation with additional rule-compliance and comfort terms and reports EPDMS. Because the two versions use different rollout and aggregation protocols, we report their results separately and do not convert one metric into the other.

For the main evaluation, the planner uses one front-camera stream and predicts an eight-point ego trajectory over a four-second horizon. The final NAVSIM v1 result is obtained on the complete navtest split containing 12,146 scenarios. The NAVSIM v2 result is evaluated with the updated official implementation and human-behavior filtering enabled.

\subsection{Candidate-Label Generation and Data Split}

The scene scorer and DAC gate are trained only on candidates retrieved for NAVSIM training scenes. Candidate labels are generated offline from the NAVSIM training metric cache using the NAVSIM/CLOVER \texttt{get\_sub\_score} evaluator. We use the batched \texttt{navsim\_v1\_style} relabeling path, with per-proposal two-way rollout disabled during label generation. The resulting labels include no-at-fault collision, drivable-area compliance, ego progress, time to collision, comfort, and the aggregate utility target used for scorer training. Final benchmark results are computed separately with the official NAVSIM v1 or v2 evaluation pipeline rather than with these training labels.

The candidate dataset contains 75,823 training scenes and 8,459 validation scenes. We use a scene-prefix split to prevent temporally adjacent samples from the same scene sequence from appearing in both subsets. The validation set and NAVSIM navtest have zero overlap in metric tokens.

\section{Scene Scorer and Feasibility Gate}

\subsection{Scene-Conditioned Scorer}

The scorer is initialized from the publicly released CLOVER trajectory scorer and re-optimized on the ground-truth-only retrieval distribution of Auto-JEPA. Its input consists of scene features, ego context, and candidate trajectory features. The optimization objective combines trajectory-component regression with within-scene ranking:
\begin{equation}
    \mathcal{L}_{\mathrm{score}}
    =\mathcal{L}_{\mathrm{comp}}
    +0.5\mathcal{L}_{\mathrm{rank}}.
\end{equation}
The ranking loss uses a temperature of 0.05 and treats candidates within 0.02 of the best target score as near-optimal. Comfort-failure candidates receive a weight of 5.0. The initial adaptation is trained for five epochs with a batch size of 32, learning rate $10^{-5}$ for the scorer modules, learning rate $10^{-4}$ for the ego adapter, and weight decay of 0.01. We then perform a three-epoch low-learning-rate continuation with learning rates $2\times10^{-6}$ and $2\times10^{-5}$, respectively. The checkpoint with the highest validation selected score is used in the final planner.

\subsection{Drivable-Area Feasibility Gate}

The feasibility gate predicts the probability that a candidate violates drivable-area constraints. It operates on the frozen candidate features used by the scorer, seven trajectory-kinematic features, and candidate-set context. The kinematic vector contains the measured ego speed, the first two candidate speeds, their initial speed mismatch relative to the ego vehicle, the absolute mismatch, and two finite-difference acceleration terms. Candidate-set self-attention allows the gate to compare each proposal with alternative motions retrieved for the same scene. The prediction trunk has a hidden dimension of 256 and dropout of 0.1.

The gate is trained with weighted binary cross-entropy and a scene-wise pairwise ranking loss,
\begin{equation}
    \mathcal{L}_{\mathrm{gate}}
    =\mathcal{L}_{\mathrm{BCE}}
    +0.3\mathcal{L}_{\mathrm{rank}},
\end{equation}
where DAC-failing candidates receive a positive-class weight of 8 and the ranking margin is 1.0. We use AdamW with a batch size of 32, learning rate $3\times10^{-4}$, weight decay $10^{-3}$, gradient clipping at 1.0, and BF16 arithmetic. During inference, candidates with predicted failure probability above $\tau=0.2$ are masked before the scorer argmax. If a scene has no remaining candidate, the system restores the ungated scorer ranking for that scene. Evaluator labels are never available to the gate during inference.

\section{Additional Ablation Details}

\subsection{Full Component Ablation}

The component ablation in the main paper isolates the roles of intent prediction, scene-conditioned scoring, and feasibility filtering. The scorer-free row still uses the DAC gate, whereas the no-gate row still uses the scene-conditioned scorer. Therefore, the comparisons should be interpreted conditionally rather than as a sequential addition of modules. The scorer contributes 3.7 PDMS when feasibility filtering is retained, while the gate contributes 0.3 PDMS and improves DAC by 0.4 points when the scorer is retained.

\subsection{Candidate-Pool Size Sensitivity}

Table~\ref{tab:supp-k-sensitivity} reports the final planning score for the candidate-pool sizes evaluated during development. Increasing the pool from one direct retrieval result to 200 candidates provides the main improvement, while increasing the pool from 200 to 300 yields a smaller additional gain.

\begin{table}[t]
    \centering
    \small
    \begin{tabular}{cr}
        \toprule
        Candidate pool size $K$ & Selected PDMS $\uparrow$ \\
        \midrule
        1   & 87.6 \\
        200 & 91.1 \\
        300 & \textbf{91.3} \\
        \bottomrule
    \end{tabular}
    \caption{Sensitivity to the number of retrieved candidates on NAVSIM v1 navtest. All settings use the same intent predictor and trajectory memory. For $K>1$, the scorer and feasibility gate select the final candidate.}
    \label{tab:supp-k-sensitivity}
\end{table}

\section{Semantic Occlusion Protocol}
\label{sec:supp-occlusion}

The semantic occlusion analysis is conducted on the complete validation split. For every valid scene, the dynamic-agent mask is formed from the projected regions of visible traffic participants and applied consistently to all four input frames. The random control masks an equal total image area. Both interventions preserve the ego-motion history and navigation command, isolating the dependence of the predicted intent on visual information.

Let $\hat{\mathbf{Z}}$ and $\hat{\mathbf{Z}}_{m}$ denote the intent representations predicted from the original and masked inputs. We measure the intervention response as
\begin{equation}
    \Delta_{\mathrm{intent}}
    =1-\cos\left(\hat{\mathbf{Z}},\hat{\mathbf{Z}}_{m}\right),
\end{equation}
where the eight temporal tokens are flattened before cosine similarity is computed. The analysis contains 15,364 valid scenes. Dynamic-agent masking produces a mean intent change of 0.080, compared with 0.027 for equal-area random masking, corresponding to a ratio of $2.97\times$. The dynamic-agent intervention produces the larger response in 71.1\% of scenes.

\section{Randomness, Runs, and Computing Infrastructure}

\subsection{Randomness Control}

We use a global seed of 42 for trajectory-space pretraining, dataset subsampling, distributed sampling, and quantitative semantic occlusion. The trajectory pretraining code applies this seed to Python, NumPy, PyTorch, and all CUDA devices. Distributed samplers receive the same base seed and use the epoch index for deterministic epoch-specific shuffling. Validation uses a fixed split and no shuffled sampler. Equal-area random masks in the semantic occlusion experiment are sampled with a NumPy generator initialized with seed 42. The qualitative figure-generation scripts use seed 2027 and enable deterministic cuDNN behavior and deterministic PyTorch algorithms when supported.

All benchmark numbers in the main paper are obtained from one deterministic full-navtest evaluation of the selected checkpoint; they are not averages over independently retrained models. The semantic occlusion statistics aggregate paired interventions over 15,364 validation samples, with the dynamic-agent and random-mask responses computed from the same checkpoint and unoccluded input. We state the number of runs explicitly because full official NAVSIM evaluation is computationally expensive and the evaluation path does not natively support multi-seed aggregation.

\subsection{Computing Infrastructure}

Training and NAVSIM evaluation were performed on Linux servers equipped with NVIDIA A100-SXM4 GPUs with 80\,GB memory. Trajectory-space pretraining and intent-predictor training support distributed execution and used one or two GPUs depending on the run; scorer adaptation, feasibility-gate training, semantic occlusion, and final benchmark evaluation used one GPU. Training commands use Python 3.12, BF16 arithmetic where stated, and one CPU thread per numerical backend during distributed runs. The released code includes the environment and dependency files needed to reconstruct the software stack. Exact package versions distributed with the release should be used rather than versions inferred from the generic upstream NAVSIM environment file.

Distributed training uses data parallelism without changing the model architecture. The per-GPU batch sizes in Table~\ref{tab:supp-hparams} therefore correspond to effective global batch sizes of 256 or 512 for trajectory pretraining and 8 or 16 for intent-predictor training when one or two GPUs are used, respectively. Scorer and gate optimization remain single-GPU procedures because their candidate features are precomputed. Final benchmark evaluation is also executed on one GPU so that all reported planner configurations use the same inference protocol.

\subsection{Complete Final Hyperparameters}

Table~\ref{tab:supp-hparams} consolidates the final settings used by the four optimized stages. The scene scorer uses the original single-head scorer attention; an eight-head continuation was evaluated during development but did not improve validation selection and is not part of the reported planner.

Frozen encoders and cached candidate features are excluded from the corresponding optimizers. Checkpoint selection uses only the validation criteria shown in the table: trajectory reconstruction uses validation ADE, the intent predictor uses the completed epoch-10 checkpoint, the scorer uses validation selected score, and the gate uses validation recall and utility. The NAVSIM navtest results are not used to choose epochs, learning rates, or thresholds.

\begin{center}
\footnotesize
\setlength{\tabcolsep}{3.5pt}
\begin{tabular}{lcc}
    \toprule
    Setting & Trajectory AE & Intent predictor \\
    \midrule
    Epochs & up to 40 & 10 total \\
    Batch size & 256/GPU & 8/GPU \\
    Optimizer & AdamW & AdamW \\
    Learning rate & $2\times10^{-4}$ & $10^{-5}$ \\
    Weight decay & 0.05 & 0.05 \\
    Dropout & 0.1 & 0.1 \\
    Gradient clipping & 1.0 & 1.0 \\
    Precision & BF16 & BF16 \\
    Selection & lowest val ADE & epoch-10 checkpoint \\
    \bottomrule
\end{tabular}

\vspace{4pt}

\begin{tabular}{lcc}
    \toprule
    Setting & Scene scorer & DAC gate \\
    \midrule
    Epochs & 5 + 3 continuation & selected checkpoint \\
    Batch size & 32 & 32 \\
    Optimizer & AdamW & AdamW \\
    Learning rate & $10^{-5}\!\rightarrow\!2\times10^{-6}$ & $3\times10^{-4}$ \\
    Auxiliary LR & $10^{-4}\!\rightarrow\!2\times10^{-5}$ & -- \\
    Weight decay & 0.01 & $10^{-3}$ \\
    Dropout & pretrained & 0.1 \\
    Gradient clipping & 1.0 & 1.0 \\
    Precision & BF16 & BF16 \\
    Selection & best val score & val recall/utility \\
    \bottomrule
\end{tabular}
\captionof{table}{Final optimization settings. ``Auxiliary LR'' denotes the ego-adapter learning rate used by the scene scorer.}
\label{tab:supp-hparams}
\end{center}

\section{Limitations and Failure Modes}

Auto-JEPA predicts the implications of a scene for future ego motion rather than rolling out a complete future environment. The learned intent is therefore an action-oriented predictive representation, not an explicit reconstruction of surrounding-agent states. This design is sufficient for the trajectory-planning interface studied here, but it does not provide the scene-level forecasts required by applications such as interactive simulation or counterfactual environment generation.

The retrieval-based planner is bounded by the coverage of its fixed trajectory memory and by the recall of the intent query. If no feasible maneuver is represented in the retrieved candidate pool, neither the scene scorer nor the feasibility gate can synthesize one. The observed saturation between $K=200$ and $K=300$ indicates that simply enlarging the candidate pool offers diminishing returns under the current memory and predictor. A broader memory, adaptive retrieval, or intent-conditioned trajectory generation and local refinement could extend the reachable motion space while preserving the learned intent representation.

When a feasible candidate is available, selection errors can still arise from scorer calibration or distribution shift. A valid candidate may receive an inaccurately low scene score, while the gate may reject a legal borderline trajectory or retain one that violates the drivable area. The all-filtered fallback guarantees a non-empty output but cannot repair an erroneous ranking. These cases motivate uncertainty-aware scoring, stronger calibration, and targeted hard-negative training.

\end{document}